# Using Deep Features of Only Objects to Describe Images

Mishra, Ashutosh
ashutosh@cs.uni-kl.de

Liwicki, Marcus
liwicki@cs.uni-kl.de

*Abstract*—Inspired by recent advances in leveraging multiple modalities in machine translation, we introduce an encoder-decoder pipeline that uses (1) specific objects within an image and their object labels (2) a language model for decoding joint embedding of object features and the object labels. Our pipeline merges prior detected objects from the image and their object labels and then learns the sequences of captions describing the particular image. The decoder model learns to extract descriptions for the image from scratch by decoding the joint representation of the object visual features and their object classes conditioned by the encoder component. The idea of the model is to concentrate only on the specific objects of the image and their labels for generating descriptions of the image rather than visual feature of the entire image. The model needs to be calibrated more by adjusting the parameters and settings to result in better accuracy and performance.

## I. Introduction

Generating descriptions for images has long been regarded as a challenging perception task integrating vision, learning and language understanding. One not only needs to correctly recognize what appears in images but also incorporate knowledge of spatial relationships and interactions between objects. Even with this information, one then needs to generate a description that is relevant and grammatically correct. With the recent advances made in deep neural networks, tasks such as object recognition and detection have made significant breakthroughs in only a short time. The task of describing images is one that now appears tractable and ripe for advancement. Being able to append large image databases with accurate descriptions for each image would significantly improve the capabilities of content-based image retrieval systems. Moreover, systems that can describe images well, could in principle, be fine-tuned to answer questions about images also. This paper describes a new approach to the problem of image caption generation, casted into the framework of encoder-decoder models. For the encoder, we learn a new embedding of object features by merging the object features of each individual object type with its class label using simple convolutions and then the decoder model tries to decode descriptions out of this new learned embedding space.

The **contribution** of our work is twofold: First, we present a new dataset for image features. The dataset provides deep features for each object instance for each image along with Global Vectors for Word Representation (GLoVe) features for each item describing the image. Secondly, we train and evaluate multiple methods for generating captions on this dataset comprising of 12000 training images from the Microsoft Common Objects in Context (MSCOCO) dataset.

The remaining sections of this paper are organized as follows: SectionII briefly describes related work in the fields of image captioning. In SectionIII, the new dataset and its unique features are illuminated. The network architecture and approach we use are presented in SectionIV. In SectionV we provide the hardware and softwares used to achieve the task. SectionVI reports the results of the experiments and SectionVII concludes the paper and gives perspectives for future work.

## II. Related Work

The task of generating natural language descriptions from visual data has long been researched and explored in the field of computer vision. Primarily the interest area was focussed more on videos for the visual data but with time, images are being seen as potentially more affective and interesting for this task. This has led to complex systems composed of visual primitive recognizers combined with a structured formal language, e.g. And-Or Graphs OR-logic systems, which are further converted to natural language via rule-based systems. Such systems worked with heavily hand-crafted features, relatively sensitive and worked for very specific fields like sports or traffic setup. Recent advances in the detection of objects, their locations in the image, their different attributes and their number, allows us to augment and generate more natural descriptions of an image. Farhadi et al. [1] use similar detections to predict triplets of scene elements which are then used for template based sentence generations. In a similar approach, Li et al. [2] apply object detections and combine together a final description using phrases containing the detected objects and learning their relationships. On the other hand Kulkarni et al. [3] apply a relatively more complex graphs of detection with template based text generation.

Works from [4]–[7] attempted to describe images "in the wild", but they mostly work on hand crafted and rigid sets of features for text generation. A large portion of research has been done in addressing the problem of ranking the description sentences of the image as well. These approaches are based on learning a common embedding vector space for both the visual as well as the textual features. For an image query, descriptions are extracted which lie close to the image in the embedding space. Neural network based learning is mostly used to learn such embedding vector spaces. The general approach is to learn visual embedding space using Convolutional

Neural Networks(CNNs) and then train a Recurrent Neural Network(RNN) to learn textual sequences. The RNN is trained to learn the context from the visual features and then associate the sequence modelling accordingly. The model is motivated by the recent success of such sequence learning ability of the RNNs in machine translation, the only differene being instead of text input for our task of image captioning we provide image visual features mostly coming from a CNN.

In a separate approach Kiros et al. [8] propose to model a joint multimodal embedding space using computer vision inspired methodology and also an LSTM for encoding the text. Such approaches have been very popular recently in handling image captioning.

## III. THE DATASET

The actual success or failure of any model is truely dependent on the type of dataset used for its training and testing. For the past twenty years the image annotation community has made great efforts in collecting rich captioned images. Our model was trained and tested only on the MSCOCO dataset which contains photos of 91 objects types that would be easily recognizable by a 4 year old.
We evaluate our method on the popular MSCOCO dataset. The dataset contains 83000 training images, 41000 validation images and 81000 images for testing. We train and test our method only on a subset of this dataset. We use 12000 training images, 6000 validation images and 1000 test images. The task associated with this dataset is to generate descriptions for the input images and to maximize the mean Bilingual Evaluation Understudy (BLEU) score. There are five written caption descriptions to each image in MSCOCO.
Based on the caption files, we created our own custom dataset to have specific details about the image as well but separately from the captions. Our custom dataset stores name of the image as an id, the number of objects in the image, labels for each of such objects, alexnet FC layter features for each instance of such objects, their bounding boxes and also their distance from the origin of the image. We believe having such resourceful dataset really augments the training pipeline by concentrating on the specific details of the image which derive the description of an image rather than just giving the whole image as an input. Figure 1 shows few sample captioned images from the MSCOCO dataset.

### A. Preprocessing

All the MSCOCO dataset used for the captioning task were primarily in JSON format as provided by the MSCOCO official website, these files needed a lot of preprocessing regarding extracting only the needed image and caption details and only for a subset of images out of the total count of 83K images. In order to accelerate the speed of training, we had created a completely new dataset based on MSCOCO dataset which contained all the relevant features of the image that our model specifically was dependent on.

## IV. METHOD

In our pursuit to design a model trained over to learn an embedding space combining visual features of all the objects and their class labels and then train a LSTM to extract novel descriptions for the image, we had tried several intermediate models and tried to learn their affects on our accuracy of the final result. Essentially we tried 3 different approaches for which detailed technical and functional ideas are mentioned below:

### A. Model 1:

- Extract image features from the fully connected layer of a pretrained CNN for example VGG16 with 4096 dimensions
- Reduce the dimensions to a lower dimensional value(128)
- Model a language component using an LSTM with 256 hidden units to learn an embedding space of size 256 for the given image image description
- Pass visual features to the LSTM with 1000 hidden units at each time step as an additional input
- Apply softmax to predict the next most probable word given the image features at each time step and also the previous word in the sequence

The idea here was to learn separate embedding spaces for each image and the ground truth image descriptions and then combine them together and then train over an LSTM to sample training descriptions for the image. Refer a similar architecture in Figure 2.

### B. Model 2:

In a separate experiment we tried a relatively complex architecture, the structure of which could be summarised as below:

- Extract image features from the fully connected layer of a pretrained CNN for example VGG16 with 2048 dimensions
- Reduce the dimensions to a lower dimensional value(128)
- Model a language component using an LSTM with 256 hidden units to learn an embedding space of size 256 for the given image image description
- Merge the two embeddings and pass it over to a bidirectional LSTM with 256 hidden units
- Apply softmax to predict the next most probable word given the image features at each time step and also the previous word in the sequence

With this approach we intended the model to learn text sequences from both the directions of the text i.e left to right(which is more natural order of reading) as well as right to left as the probability of occurance of a particular word in a sentence is highly correlated with the words appearing both in the left and right neighbourhood.

### C. Model 3:

The third approach we tried is a novel architecture where we extract features separately from object visual embeddings from alexnet and object labels from GLOVE embedding. The extracted features after convolutions are merged together by applying a concatenation operation and then fed as input at each time step to an LSTM which works same as the previous models to learn the text description sequences for the given image. Refer the figure 3 for the model architecture.

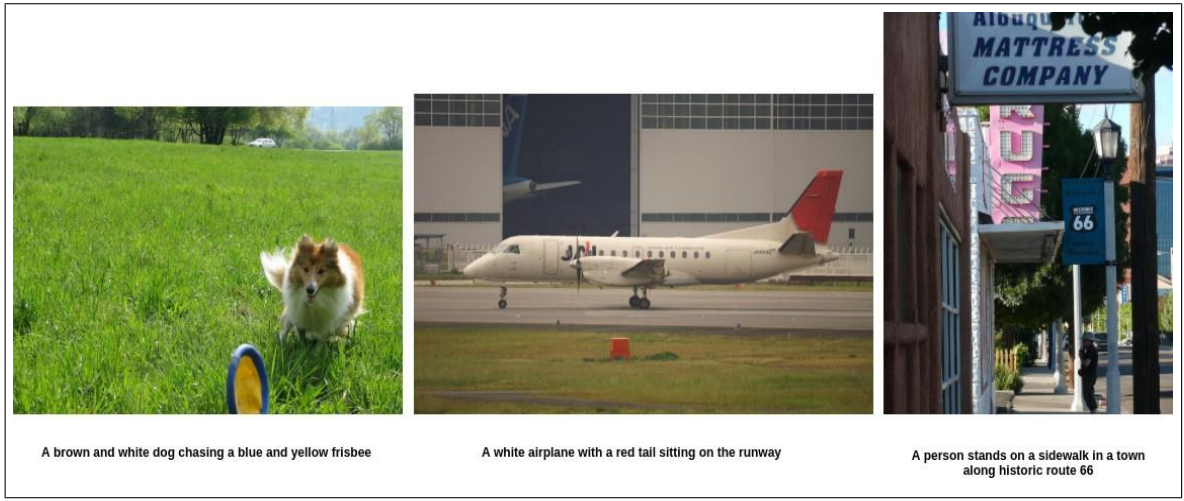

Fig. 1: Sample images from MSCOCO dataset

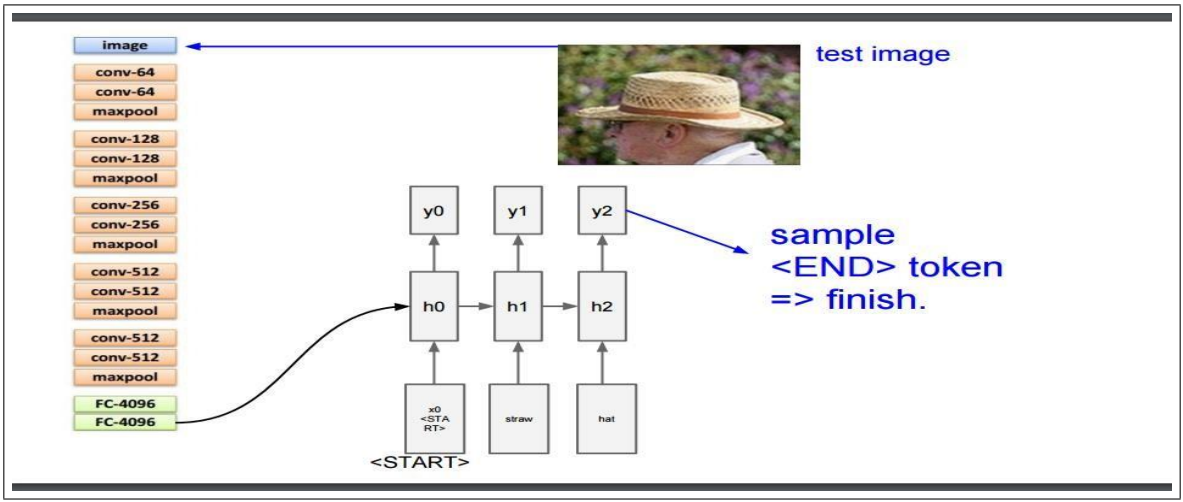

Fig. 2: Basic network architecture

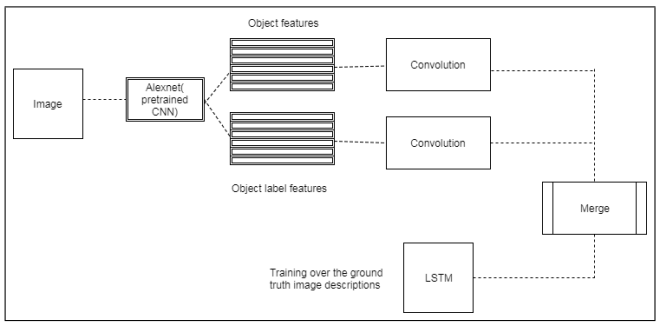

Fig. 3: Network architecture for our model

## V. HARDWARE AND SOFTWARE DETAILS

All processing was carried out on commodity hardware, specifically a 64bit Intel i7 processor (32GB ram, Graphic TITAN X(Pascal)/PCLe/SSE2) running an Ubuntu 14.04 LTS operating system. We used Python 2.7 for programming as it is a mature, versatile and robust high level programming language. It is an interpreted language which makes the testing and debugging phases extremely quick as there is no compilation step. There are extensive open source libraries available for this version of python and a large community of users.

## VI. RESULTS

In the following, we report the results of the experiments and attempt to give reasons for the outcomes.

Since our model is data driven and trained end-to-end, and given the abundance of datasets, we wanted to answer questions such as how dataset size affects generalization and "How number of iterations of same images affects the learning process". In order to test these ideas, we trained our Model 1 on 6000 images from MSCOCO for 15 epochs and tested the same over 1000 images from the same dataset. In a separate experiment we trained our Model 2 over 1500 images with 3 and 15 epochs and Model 3 with 6000 images separately with 15 epochs. Refer Figure 4, 5 and 6 for the relevant test results.

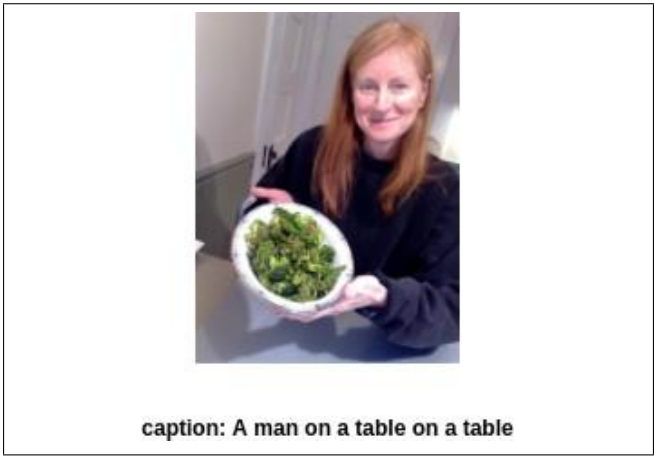

Fig. 4: Test result for Model 1 trained with 6000 training images and 15 epochs

TABLE I: Comparison with State of the art results

|  | Description | BLEU |
|---|---|---|
| TencentVision | Multi-attention and RL | 0.795 |
| panderson@MSR/ACRV | Bottom-Up and Top-Down Attention | 0.802 |
| DEEPAI | dascup | 0.786 |
| our model(3) | With Deep Object only Features | 0.126 |

## VII. CONCLUSION AND FUTURE WORK

We believe that by combining specific object features with their label vector embeddings does not decrease the accuracy of an image captioning model which seemed probable at the start. We still are getting pretty much the same results with a BLEU score of 0.126. We believe, we can experiment more exhaustively with the object feature representation and label features in order to improve the results. As a future work, we propose to study if weighing specific objects features based on their location in the image and their size affects the results or not. As part of our experiments it can be safely concluded that for our image captioning models images of the order ¿ 1500 and epochs of the order of 15 are necessary to train the models, as in our experiment training with just 1500 images resulted very poorly with only 0.014 BLEU score and increasing the epochs also resulted in a relatively better performance.

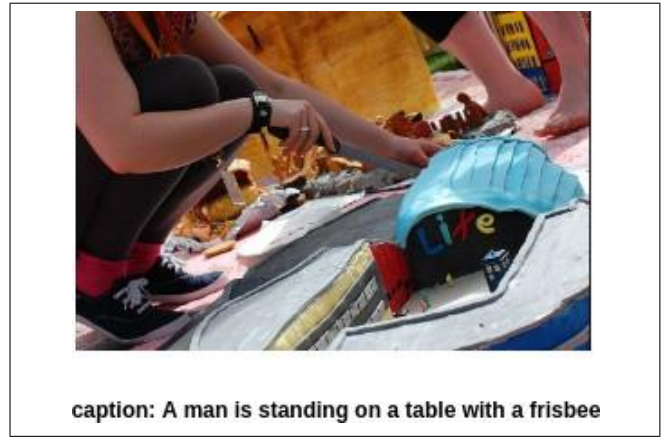
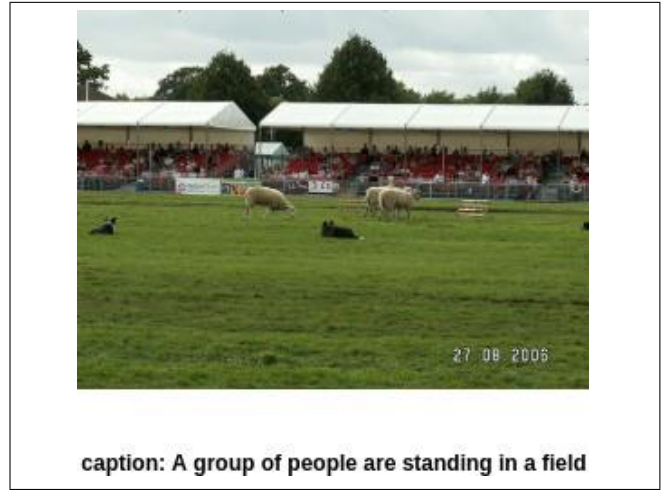

Fig. 5: Test result for Model 2 trained with 6000 training images after 3 and 15 epochs

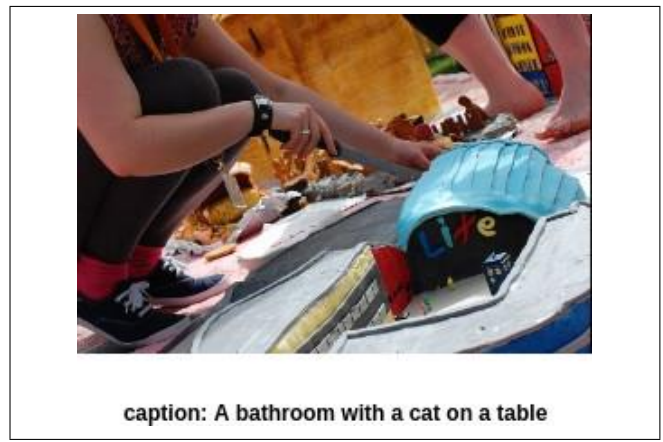

Fig. 6: Test result for Model 3 trained with 6000 training images and 15 epochs